\documentclass[11pt]{article}

\usepackage[letterpaper, text={6.5in, 9.0in}]{geometry}
\usepackage{amsmath}
\usepackage{amssymb}
\usepackage{amsthm}
\usepackage{natbib}
\usepackage{graphicx}
\theoremstyle{definition}

\makeatother

\newtheorem{Proposition}{Proposition}
\newtheorem{corollary}{Corollary}

\def\bw{\mbox{\boldmath $\theta$}}
\def\bw{\mbox{\boldmath $\Theta$}}
\def\bSigma{\mbox{\boldmath $\Sigma$}}

\def\bmu{\mbox{\boldmath $\mu$}}

\def\bz{{\bf z}}

\def\bX{{\bf X}}
\def\bR{{\bf R}}

\def\bQ{{\bf Q}}

\def\bI{{\bf I}}

\def\bw{{\bf w}}

\def\bu{{\bf u}}
\def\bv{{\bf v}}
\def\bx{{\bf x}}
\def\by{{\bf y}}

\def\mR{\mathbb{R}}
\def\bR{\mathbb{R}}
\def\bE{\mathbb{E}}
\def\nn{\nonumber}

\def\v2{\vspace{0.2in}}

\providecommand{\keywords}[1]
{
  \textbf{\textit{Keywords---}} #1
} 

\author{Hanwen Huang \\
    {\it Department of Biostatistics, Data Science and Epidemiology}\\
    {\it Medical College of Georgia, Augusta University, Augusta, 30912}\\
    hhuang1@augusta.edu\\\\
Peng Zeng \\
{\it Department of Mathematics \& Statistics}\\
{\it Auburn University, Auburn, AL 36849}\\
zengpen@auburn.edu}

\begin{document}
\title{Statistical Inference in Classification of High-dimensional Gaussian Mixture}
\maketitle

\begin{abstract}
We consider the classification problem of a high-dimensional mixture of two Gaussians with general covariance matrices. Using the replica method from statistical physics, we investigate the asymptotic behavior of a general class of regularized convex classifiers in the high-dimensional limit, where both the sample size $n$ and the dimension $p$ approach infinity while their ratio $\alpha=n/p$ remains fixed. Our focus is on the generalization error and variable selection properties of the estimators. Specifically, based on the distributional limit of the classifier, we construct a de-biased estimator to perform variable selection through an appropriate hypothesis testing procedure. Using $L_1$-regularized logistic regression as an example, we conducted extensive computational experiments to confirm that our analytical findings are consistent with numerical simulations in finite-sized systems. We also explore the influence of the covariance structure on the performance of the de-biased estimator.
\end{abstract}

{\keywords{
De-biased estimator; Generalization error; Logistic regression; $L_1$-regularization; Variable selection.
}}

\section{Introduction}
\label{section:introduction}
It is well known that, due to the development of computer technology, modern statistical problems are increasingly high-dimensional, i.e. the number of parameters $p$ is large. Examples abound from genetic molecular measurements (where many gene level features have been measured), chemometrics (high-dimensional spectra), medical image analysis (3-d shapes represented by high-dimensional vectors), and so on. On the other hand, given the advent of a large volume of data collected through various sources, especially the Internet, it is also important that we consider the case where the sample size $n$ is large. Thus, developing efficient inference tools to tackle challenges in large-scale data where both the sample size $n$ and the dimension $p$ are huge is a crucial task in statistical modeling. However, many existing statistical theory and methods that were developed to solve small data problems have difficulty being applied to large-scale data problems.

In this paper, we focus on classification methods that fit within a general regularization framework, formulated as the minimization of certain penalized loss functions. This framework encompasses many commonly used classification methods, such as logistic regression and support vector machines (SVM), as special cases. The statistical inference of this class of methods in low dimensions has been well studied in the literature; see, e.g., \cite{svminf,JMLR:v20:18-801}. To deal with high-dimensional problems where the number of parameters is larger than the number of samples, people usually use a sparse regularization approach such as $L_1$-penalty. However, the use of sparse regularization comes at a price because usually it is impossible to characterize the distributions of the sparse estimators. Thus, it is quite challenging to construct an inference procedure to quantify the uncertainty of the estimated parameters. This is different from the classical statistics, where the exact distributions are available, or the approximated distributions can be derived from a large-sample asymptotic theorem.

Recent rapid advances in statistical theory about the asymptotic performance of many classic machine learning algorithms in the limits of both large $n$ and $p$ have shed some light on this issue. There has been considerable effort to establish asymptotic results for regularized convex classification methods under the assumption that $p$ and $n$ grow at the same rate $n/p\rightarrow\alpha > 0$. This is different from the traditional large sample theory in statistics, which deals with asymptotic behavior as $n\rightarrow\infty$ with $p$ fixed. The advantage of this type of asymptotics is that it can provide sharp quantitative guidelines on how to design machine learning systems when both $p$ and $n$ are large but finite numbers.

\cite{Huang17,maistatistical} studied SVM under Gaussian mixture models in which data is assumed to be generated from a Gaussian mixture distribution with two components, one for each class. The covariance matrix is assumed to follow a spiked population model. In the same setting,  \cite{mailiao,huang2019large} studied regularized logistic regression and general margin-based classification methods, respectively. \cite{montanari2019generalization} studied the hard margin SVM under the single Gaussian model in which the data are assumed to be generated from a single Gaussian distribution. \cite{dengmodel} studied unregularized logistic regression under two-component Gaussian mixture models. The sharp asymptotics for the unregularized logistic regression have been studied in \cite{candes2020phase} under the single Gaussian models. \cite{gerace2020generalisation} studied the classification error for $L_2$ regularized logistic regression for a single Gaussian model with a two-layer neural network covariance structure. Analogous results for Gaussian mixture models with standard Gaussian components were provided in \cite{mignacco2020role}. The sharp asymptotics of generic convex generalized linear models were studied in \cite{gerbelot2020asymptotic} for rotationally invariant Gaussian data and in \cite{loureiro2021capturing} for block-correlated Gaussian data. The multiclass classification for the Gaussian mixture was also provided in \cite{loureiro2021learning} recently.  

Although most of the results in the aforementioned literature focus on investigating the generalization error of classifiers, the influence of regularization effects on variable selection has remained largely unexplored. The goal of this paper is to construct statistical inference procedures for the $L_1$-regularized convex classification methods in high-dimensional settings, that is, to calculate confidence intervals and p-values for parameters estimated from the model. Our results are based on the asymptotic behavior of the estimators in the limit of both $n\rightarrow\infty$ and $p\rightarrow\infty$ at a fixed rate $n/p\rightarrow\alpha$ under two-component Gaussian mixture models. We derive the analytical results using the replica method developed in statistical mechanics. All analytical results are confirmed by numerical experiments on finite-size systems and thus our formulas are verified to be correct.

Note that some of the results in the aforementioned literature are rigorous under Gaussian assumption. The rigorous analysis methods include convex random geometry \citep{candes2020phase}, random matrix theorem \citep{dobriban2018high}, message-passing algorithms \citep{bayati2011lasso,berthier2020state,loureiro2021learning}, convex Gaussian min-max theorem \citep{montanari2019generalization,mignacco2020role,dengmodel}, and interpolation techniques \citep{barbier2018adaptive}. The rigorous work
so far mainly focuses on `i.i.d. randomness', corresponding to the case of standard Gaussian
design. The present work focuses on the mixture of two Gaussian components with arbitrary covariance structure. While it remains an open problem to derive a rigorous proof for our results, we shall use simulation on moderate system sizes to provide numerical support that the theoretical formula is indeed exact in the high-dimensional limit.

The remainder of this paper is organized as follows. In Section \ref{section:method}, we state the asymptotic results in the joint limit of large $p$ and $n$ for general $L_1$ regularized convex classification methods. Based on these asymptotic results, in Section \ref{section:implementation}, we develop the computing algorithm for implementing the corresponding statistical inference procedures. In Section \ref{numeric}, using $L_1$-regularized logistic regression as an example, we present numerical studies by comparing the theoretical results with Monte Carlo simulations on finite-size systems under different experimental settings. The last section is devoted to the conclusion, and the derivation of the main analytical results is provided in the Appendix.

\section{De-biased estimator of penalized classification}\label{section:method}

In binary classification, we are typically given a random sample $\{(\bx_i,y_i)\}^n_{i=1}$, where $y_i\in\{1,-1\}$ denotes the categorical label and $\bx_i\in{\mR}^p$ denotes the input covariates. The goal of linear classification is to estimate a vector $\bw\in{\mR}^p$ such that sign($\bx^T\bw$) can be used to predict output labels for future observations with input covariates $\bx$ only. The focus of this paper is on the penalized convex classification methods that can be fit in the following $loss$ + $penalty$ regularization framework
\begin{eqnarray}\label{class}
\hat{\bw}&=&\arg\min_{\bw\in\mR^p}\left\{\sum_{i=1}^nV\left(\frac{y_i\bx_i^T\bw}{\sqrt{p}}\right)+\sum_{j=1}^pJ_\lambda(w_j)\right\},
\end{eqnarray}
where $V(u)$ is the convex loss function and $J_\lambda(w)=\lambda|w|$ is the $L_1$ regularization term. 
The general requirement for the loss function is convex, decreasing, and $V(u)\rightarrow 0$ as $u\rightarrow\infty$. Many commonly used classification techniques can be fit into this regularization framework. Examples include penalized logistic regression (PLR; \cite{lin2000}) and support vector machine (SVM; \cite{Vapnik95}). The loss functions of these classification methods are
\begin{eqnarray}\nn
\text{PLR}:&&V(u)=\log[1+\exp(-u)],\\\nn
\text{SVM}:&&V(u)=(1-u)_+.
\end{eqnarray}
In addition to the above methods, many other classification techniques can also fit the regularization framework, for example, distance-weighted discrimination (DWD; \cite{Marron2007}), the unified machine with large margin \citep{liu2011}, the AdaBoost in Boosting \citep{FREUND1997119,friedman2000}, the import vector machine (IVM; \cite{ivm}), and $\psi$ learning \citep{shen2003}. 

We use the replica method to study the asymptotic behavior of the classification method (\ref{class}) in the large system limit of $n,p\rightarrow\infty$ with fixed ratio $\alpha=n/p$. This type of asymptotics is very useful because it can provide not only bounds, but also sharp predictions for the limit of the joint distribution of the estimated and true parameters in a model. From the joint distribution, various further computations can be made to provide precise predictions of quantities such as the mean squared error and the misclassification error rate. The replica method is a technique developed in statistical physics to analyze the properties of disordered systems in the limit of infinite system size. It is an appropriate tool for the theoretical analysis of high-dimensional problems. Although it is a heuristic method and some of its steps involve mathematical subtleties \citep{mezard2009information}, it has been applied successfully in a number of very difficult problems in a variety of scenarios \citep{PhysRevLett.43.1754,talagrand2003spin}. 

Assume that each training data $(\bx_i,y_i)$ for $i=1,\cdots,n$ is an independent random vector distributed according to a joint distribution function $p(\bx,y)$.  Conditional on $y=+1,-1$, $\bx$ follows multivariate normal distributions $p(\bx|y=+1)$, $p(\bx|y=-1)$ with mean $\bmu_+,\bmu_-$ and covariance $\bSigma_+,\bSigma_-$, respectively. Without loss of generality, assume $\bmu_+=-\bmu_-=\bmu$. Here $\bmu\in\mR^p$ and $\bSigma_{\pm}$ are $p\times p$ positive definite matrices. In this setting, data are generated from a mixture of two multivariate Gaussian distributions with different means and covariance matrices. The size of the signal can be characterized by $\mu=\|\bmu\|$. Here, $\bSigma_{\pm}$ can be any finite positive definite matrix which makes our model quite general. For vectors $\bu,\bv\in\mR^p$, denote $\|\bu\|^2_{\bSigma}=\bu^T\bSigma\bu$ and $\langle\bu,\bv\rangle=\sum_{j=1}^pu_jv_j$. Assuming $\bSigma_+=\bSigma_-=\bSigma$, the following proposition characterizes the limiting distribution of the solution $\hat{\bw}$ to (\ref{class}).
\begin{Proposition}\label{prop:main}
Define two random vectors
\begin{eqnarray}\label{df1}
\bar{\bw}&=&\hat{\bw}-\frac{1}{\sqrt{p}\zeta}\sum_{i=1}^ny_iV^\prime\left(\frac{y_i\bx_i^T\hat{\bw}}{\sqrt{p}}\right)\bSigma^{-1}\bx_i,\\\label{df2}
\tilde{\bw}&=&\sqrt{p}R_0\bSigma^{-1}\hat{\bmu}/\zeta+\tau\bSigma^{-1/2}\bz,
\end{eqnarray}
where $\hat{\boldsymbol{\mu}} = \boldsymbol{\mu}/\|\boldsymbol{\mu}\|$, $\hat{\bw}$ is the minimizer of (\ref{class}), $\bz\sim N(0,\bI_{p\times p})$, $\tau=\sqrt{\zeta_0}/\zeta$, and $\zeta_0,\zeta,R_0$ can be solved from the following set of nonlinear equations:
\begin{eqnarray}\nn
   \zeta_{0} &=& \frac{\alpha}{q^{2}}\bE(\hat u_\epsilon - R\mu - \sqrt{q_{0}}\epsilon)^{2},\\\nn
   \zeta&= &-\frac{\alpha}{q\sqrt{q_{0}}}\bE[(\hat u_\epsilon - R\mu - \sqrt{q_{0}}\epsilon)\epsilon],\\\label{nle}
   R_{0}&= &\frac{\alpha\mu}{q}\bE(\hat u_\epsilon - R\mu - \sqrt{q_{0}}\epsilon),\\\nn
    q_0&=&\frac{1}{p}\bE(\hat{\bw}_{\bz}^T\bSigma\hat{\bw}_{\bz}),\\\nn
    q&=&\frac{1}{p\sqrt{\zeta_0}}\bE(\hat{\bw}_{\bz}^T \bSigma^{1/2} \bz),  \\\nn
    R&=&\frac{1}{\sqrt{p}}\bE(\hat{\bw}_{\bz}^T\hat{\bmu}),
\end{eqnarray}
where the first three expectations are with respect to $\epsilon\sim N(0,1)$, the last three expectations are with respect to $\bz\sim N(0,\bI_{p\times p})$, and
\begin{eqnarray}\label{hatu}
\hat{u}_\epsilon&=&\text{argmin}_{u\in \mR}\left[V(u)+\frac{(u-R\mu-\sqrt{q_0}\epsilon)^2}{2q}\right],\\\label{hatz}
\hat{\bw}_{\bz}&=&\text{argmin}_{\bw\in \mR^p}\left[\frac{\zeta}{2}\|\bw\|_{\bSigma}^2-\langle\sqrt{\zeta}_0\bSigma^{1/2}\bz+\sqrt{p}R_0\hat{\bmu},\bw\rangle+\sum_{j=1}^pJ_\lambda(w_j)\right].
\end{eqnarray}
Then $\bar{\bw}$ and $\tilde{\bw}$ follow the same distribution asymptotically in the limit of $n,p\rightarrow\infty$ with fixed $\alpha=n/p$.
\end{Proposition}
The derivation of Proposition \ref{prop:main} is given in the Appendix based on the replica method. The two random vectors defined in (\ref{df1}) and (\ref{df2}) can be considered as the modified estimator and the signal, respectively. This is analogous to the de-biased estimator in LASSO, where $\bar{\bw}$ can be interpreted as the de-biased estimator of the penalized convex classification method (\ref{class}), providing a foundation for statistical inference. The random vector $\tilde{\bw}$ consists of two terms: the first corresponds to the true signal $\bmu$, while the second represents noise related to high-dimensional effects. It is important to note that this paper focuses on theoretical analysis. In practice, 
$\bSigma$ is typically unknown and must be estimated. The distributional limit of the estimator $\hat{\bw}$ as derived in Proposition \ref{prop:main}, enables us to calculate the prediction accuracy of the classifier, as detailed in the following corollary.
\begin{corollary}\label{cor}
Under the conditions of Proposition \ref{prop:main}, the limiting distribution of $\hat{\bw}$ leads to the asymptotic precision 
\begin{eqnarray}\label{prec}
P\{y\bx^T\hat{\bw}\ge 0\}\rightarrow\Phi\left(\frac{R\mu}{\sqrt{q_0}}\right),
\end{eqnarray}
where the parameters $R,q_0$ are determined from the set of nonlinear equations (\ref{nle}) and $\Phi$ is the CDF of the standard normal distribution.
\end{corollary}

Toward variable selection for the penalized convex classification method (\ref{class}), we consider the population loss $\bE[V(y\bx^T\bw)]$. Let $\bw_0\in\mR^p$ denote the true parameter value, which is defined as the minimizer of the population loss, i.e. $\bw_0=\text{argmin}_{\bw\in\mR^p}{\bE}[ V(y\bx^T\bw)]$. This definition has a strong connection to the Bayes rule, which is theoretically optimal if the underlying distribution is known. The Bayes rule is given by $\text{sign}(\bx^T\bw_\text{Bayes})$ with $\bw_\text{Bayes}=\text{argmin}_{\bw\in\mR^p}{\bE}[I\{\text{sign}(\bx^T\bw)\ne y\}]$. The Bayes rule is unattainable if we assume that we have no knowledge of the high-dimensional conditional density $p(\bx|y)$. Note that $\bw_\text{Bayes}$ and $\bw_0$ are equivalent to each other in the important special case of Fisher linear discriminant analysis. In more general settings, $\bw_\text{Bayes}$ and $\bw_0$ may not be the same. The minimizer from the population loss could be a reasonable target for inference in many applications \citep{wuwang,liulu}. 

If the conditional distribution $p(\bx|y)$ is multivariate normal, $\bw_0$ is proportional to $\bSigma^{-1}\hat{\bmu}$. In general settings, $\bw_0$ is approximately proportional to $\bSigma^{-1}\hat{\bmu}$ which is in the same direction as the first term on the right-hand side of (\ref{df2}). The basic intuition of this observation is that, according to equations (\ref{df1}) and (\ref{df2}), the marginal distribution of $\bar{w}_j-cw_{0,j}$ is expected to be asymptotically $N(0,\tau^2(\bSigma^{-1})_{jj})$ for $j=1,\cdots,p$, where $c$ is a constant. Assume that the true parameter $\bw_0$ is sparse, then variable selection can be achieved using a series of hypothesis testing procedures. Here, we consider testing the null hypothesis $H_{0j}: w_{0,j}=0$ for $j=1,\cdots,p$ and assigning the $p$-values for these tests. Rejecting $H_{0j}$ is equivalent to stating that $w_{0j}\ne 0$. The decision rule for $j$-th hypothesis can be based on the $p$-value $p_j=2\left(1-\Phi\left(\left|\bar{w}_j/\tau/\sqrt{\left(\bSigma^{-1}\right)_{jj}}\right|\right)\right)$, where $\Phi(x)$ is the standard Gaussian distribution function. The confidence interval for the $j$-th coefficient can be estimated as 
\begin{eqnarray}\label{confi}
\left[\bar{w}_j-\Phi(1-\alpha/2)\tau\sqrt{\left(\bSigma^{-1}\right)_{jj}}, \; 
\bar{w}_j+\Phi(1-\alpha/2)\tau\sqrt{\left(\bSigma^{-1}\right)_{jj}}\right], 
\end{eqnarray}
where $\alpha$ is the prespecified significance level, not to be confused with the ratio of $n/p$. 

\section{Implementation}
\label{section:implementation}

The parameters $\zeta_0$, $\zeta$, $R_0$, $q_0$, $q$, and $R$, which are determined by the nonlinear equations in (\ref{nle}) of Proposition~\ref{prop:main}, play an important role in characterizing the properties of $\hat{\bw}$. This section discusses the algorithm used to calculate these parameters.

For the $L_1$-penalty, (\ref{nle}) does not have a closed-form solution, so we use an iterative numerical algorithm to solve it. The parameters can be partitioned into two groups: $\{\zeta_0, \zeta, R_0\}$ and $\{q_0, q, R\}$, with one group updated while the other is held fixed in each iteration. Starting with an initial set of values $\{\zeta_0^{(0)}, \zeta^{(0)}, R_0^{(0)}\}$, we iteratively compute these parameters as follows for $k = 1, 2, \ldots$ until convergence. 

\begin{itemize}
\item 
For given $\{\zeta_0^{(k)}, \zeta^{(k)}, R_0^{(k)}\}$, calculate $\{q_0^{(k+1)}, q^{(k+1)}, R^{(k+1)}\}$ by
\begin{align*}
    q_0^{(k+1)} &= \frac{1}{p}\bE(\hat{\bw}_{\bz}^T\bSigma\hat{\bw}_{\bz}), \\
    q^{(k+1)} &= \frac{1}{p\sqrt{\zeta_0^{(k)}}}\bE(\hat{\bw}_{\bz}^T \bSigma^{1/2} \bz),  \\
    R^{(k+1)} &= \frac{1}{\sqrt{p}}\bE(\hat{\bw}_{\bz}^T\hat{\bmu}),
\end{align*}
where $\hat{\bw}_{\bz}$ is given in (\ref{hatz}) which depends on $\{\zeta_0^{(k)}, \zeta^{(k)}, R_0^{(k)}\}$. It requires high-dimensional numerical integration to compute these expectations because there is no explicit expression for $\hat{\bw}_{\bz}$ when $J_\lambda(w)=\lambda|w|$ and $\bSigma$ are general. 
In our implementation, we use Monte Carlo integration to compute the expectation. For a general $\bSigma$, (\ref{hatz}) is essentially a LASSO problem that can be efficiently solved by the coordinate descent algorithm \citep{Wu_2008}. 

\item 
For given $\{q_0^{(k+1)}, q^{(k+1)}, R^{(k+1)}\}$, calculate $\{\zeta_0^{(k+1)}, \zeta^{(k+1)}, R_0^{(k+1)}\}$ by 
\begin{align*}
   \zeta_{0}^{(k+1)} 
   &= \frac{\alpha}{{q^{(k+1)}}^{2}} 
      \bE\left(\hat u_\epsilon - R^{(k+1)}\mu - \sqrt{q_{0}^{(k+1)}}\epsilon\right)^{2}, \\
   \zeta^{(k+1)}
   &= -\frac{\alpha}{q^{(k+1)}\sqrt{q_{0}^{(k+1)}}} 
      \bE\left[\left(\hat u_\epsilon - R^{(k+1)}\mu - \sqrt{q_{0}^{(k+1)}}\epsilon\right)\epsilon\right], \\
   R_{0}^{(k+1)} 
   &= \frac{\alpha\mu}{q^{(k+1)}} 
      \bE\left(\hat u_\epsilon - R^{(k+1)}\mu - \sqrt{q_{0}^{(k+1)}}\epsilon\right), 
\end{align*}
where $\hat{u}_{\epsilon}$ is given in (\ref{hatu}). The crucial step in this iteration involves evaluating the integrations $\bE(\hat u_{\epsilon}^2)$, $\bE(\hat u_{\epsilon}\epsilon)$, and $\bE(\hat u_{\epsilon})$. Since there is no explicit expression for $\hat u_{\epsilon}$ in many applications, these expectations are computed numerically. Because $\hat u_{\epsilon}$ is univariate, numerical integration is shown to be efficient and straightforward. Taken logistic regression as an example which is considered one of the most popular and successful classification methods available on the shelf, $\hat u_{\epsilon}$ can be obtained as the the root of equation
\begin{align*}
    \frac{1}{e^u + 1} = \frac{u - R^{(k+1)}\mu - \sqrt{q_0^{(k+1)}}\epsilon}{q^{(k+1)}},
\end{align*}
which is obtained by setting the first-order derivative of the objective function in (\ref{hatu}) to be zero. 

\end{itemize}

\section{Simulation}\label{numeric}

This section presents simulation studies for $L_1$ regularized logistic regression (PLR) to validate the findings discussed in the previous sections. Unlike the existing literature, which usually assumes low correlation among predictors, we generate synthetic data including both low correlation and high correlation among predictors. The $L_1$-regularized logistic regression model is fitted using the R package \verb"glmnet". 


\subsection{Precision rates}
\label{example:precision}

This subsection compares the empirical precision rates computed by Monte Carlo and those computed from the asymptotic result based on Corollary \ref{cor}. 
We set $p = 1000$ and $n = p\alpha$, where $\alpha = 0.5$. The response variable $y$ takes values in $\{-1, +1\}$ with equal probabilities for each. 
The predictors $\bx\mid y$ are simulated from $N_p(y\bmu, \bSigma)$ for various configurations of $\bmu$ and $\bSigma$. 
We choose $\bmu = a \bSigma \bw_0$, where $\bw_0$ is a sparse vector with components equal to 1 or 0 with sparsity level $\epsilon = 0.01, 0.05,0.1$, and $a$ is selected to ensure that the length of $\bmu$ is 2. 
Set $\bSigma = \sigma^2 \bR$, where $\sigma^2 = 2.0$. We consider four different correlation structures for $\bR$: 
(1) IID, $\bR = \bI_p$, the identity matrix; 
(2) block diagonal, $\bR$ is a block diagonal matrix with blocks $\begin{pmatrix}1 & 0.8 \\ 0.8 & 1 \end{pmatrix}$;
(3) AR1, $\bR = (0.8^{|i-j|})$ representing an autoregressive model of order 1;
(4) banded, $\bR$ is a banded correlation matrix where the diagonal elements are $R_{ii} = 1$ and the off-diagonal elements are $R_{ij} = 0.4$ if $i\neq j$ and $|i - j| \leq 2$ and $R_{ij} = 0$ otherwise.

The precision rate of a classifier is the probability of correctly classifying a new observation. Let $\hat\bw$ be the fitted coefficient from a simulated dataset. The corresponding precision rate is defined as $P(\tilde y \tilde \bx^T \hat\bw > 0)$ for a new observation $(\tilde y, \tilde \bx)$. In practice, we randomly simulate $(\tilde y, \tilde \bx)$ using the same mechanism as the original training data of the same size and approximate the accurate probability of classification by a proportion of samples. 
As $n, p\to\infty$, the asymptotic precision rate is given by $\Phi(R\|\bmu\|/\sqrt{q_0})$ in (\ref{prec}), where $\Phi(\cdot)$ is the CDF of the standard normal distribution, $R$ and $q_0$ are saddle point parameters evaluated according to the algorithm described in Section~\ref{section:implementation}. 

\begin{figure}[hb]
\begin{center}
\includegraphics[width = 0.48\textwidth]{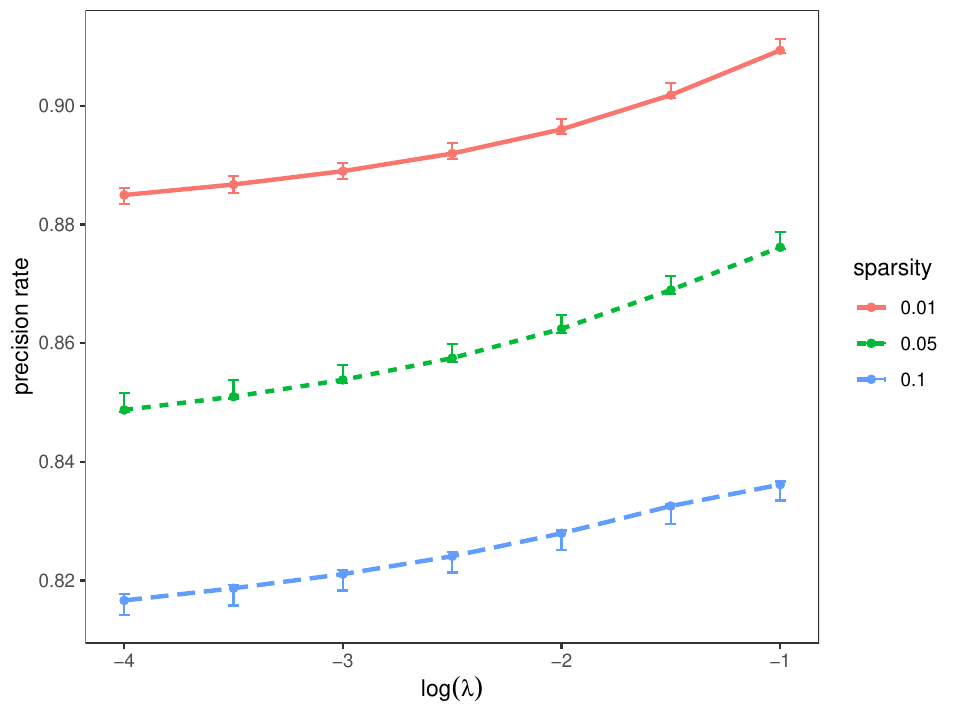}
\includegraphics[width = 0.48\textwidth]{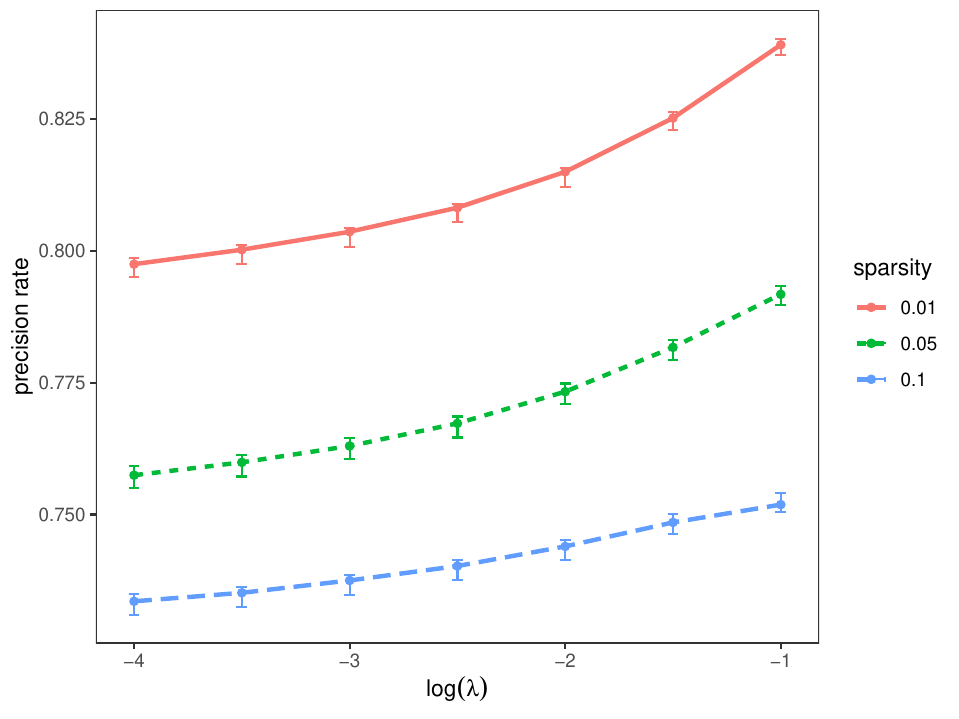}
\includegraphics[width = 0.48\textwidth]{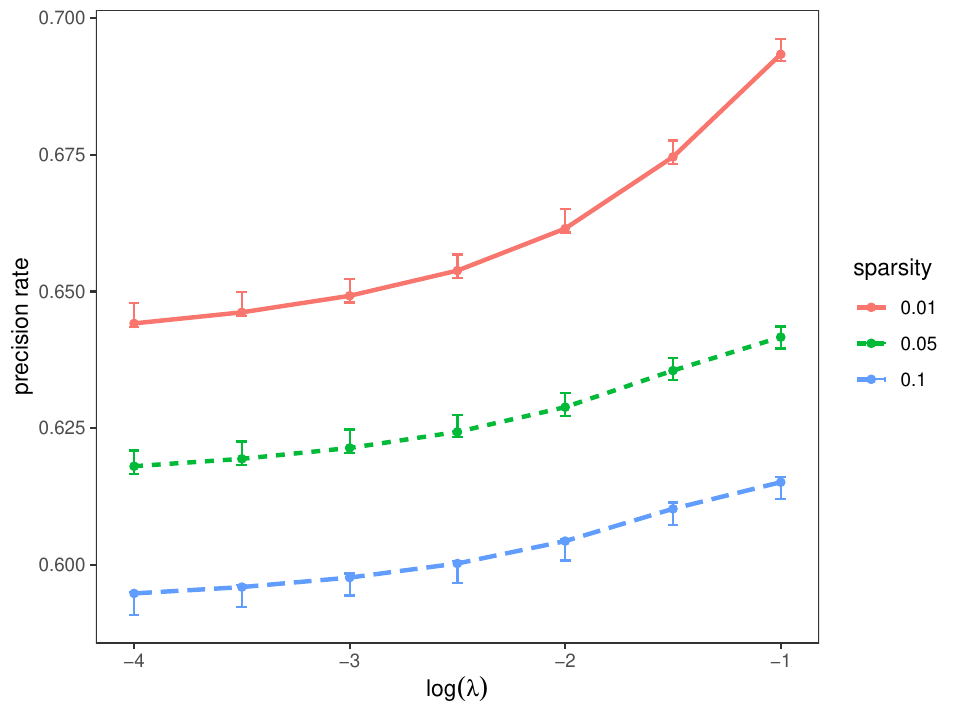}
\includegraphics[width = 0.48\textwidth]{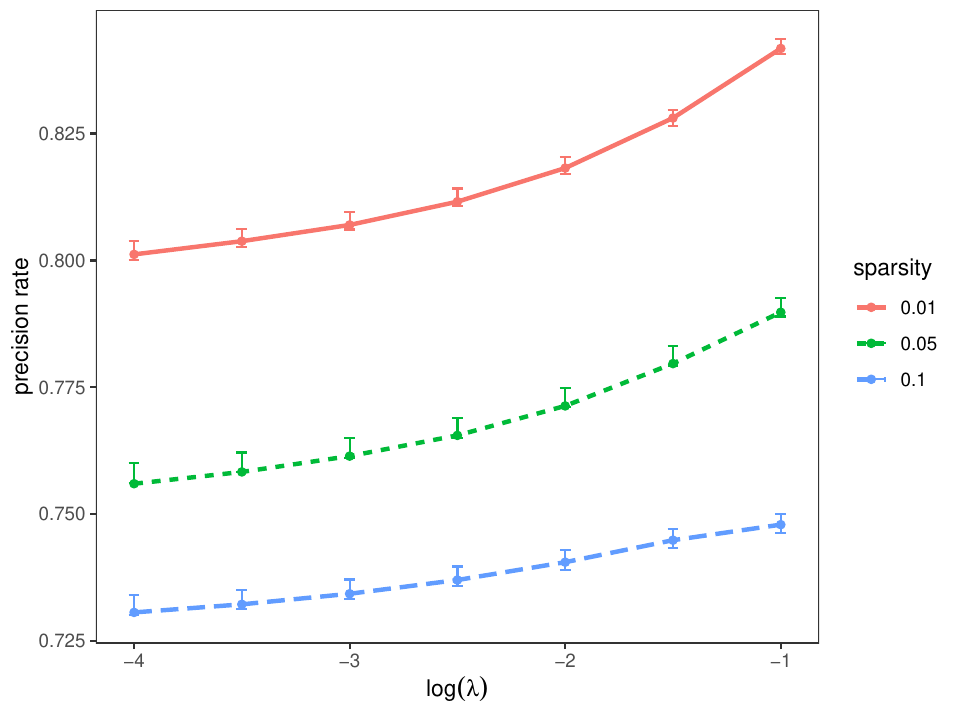}
\end{center} 
\caption{Comparison between theoretical and empirical precision rates for four different correlation structures: IID (top-left), block (top-right), AR1 (bottom-left), and banded (bottom-right). In each plot, the three lines are the theoretical precision rates at different sparsity levels $\epsilon = 0.01, 0.05, 0.1$. The error bars are the 95\% confidence intervals of the mean precision rates based on 500 replicates.}
\label{fig:precision}
\end{figure}

Figure~\ref{fig:precision} presents results from 500 replicates for different correlation structures and sparsity levels. The model is fitted using various tuning parameters $\log \lambda = -4, -3.5, \ldots, -1$. The three lines represent the precision rates computed from the asymptotic results at different sparsity levels, while the error bars denote the confidence intervals of the mean precision rates calculated from Monte Carlo. The figure illustrates a close alignment between the empirical findings and the theoretical predictions. The precision increases with sparsity for fixed covariance structure. Under the same sparsity level, the precision decreases as the covariance structure becomes more complicated. Specifically, the IID case gives the highest precision, and the AR1 case gives the lowest precision. The precision rates for the block diagonal and banded cases are in-between. Plots for other configurations exhibit similar consistency which are not shown here due to space limitation.

\subsection{Variable selection}
\label{example:CI}

\begin{figure}[hb]
\begin{center}
\includegraphics[width = 0.48\textwidth]{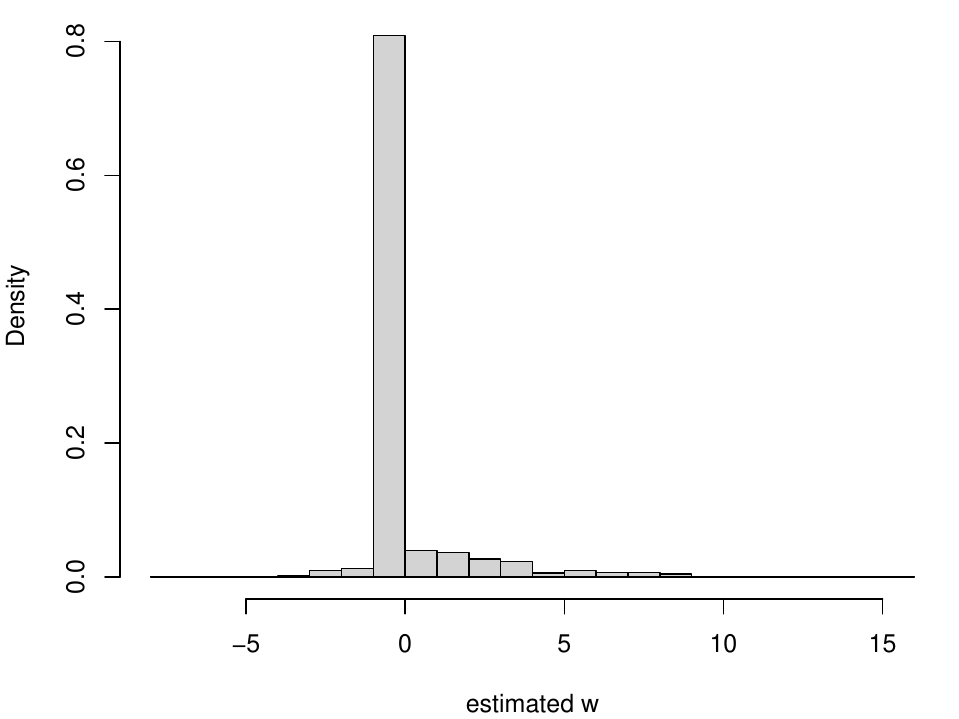}
\includegraphics[width = 0.48\textwidth]{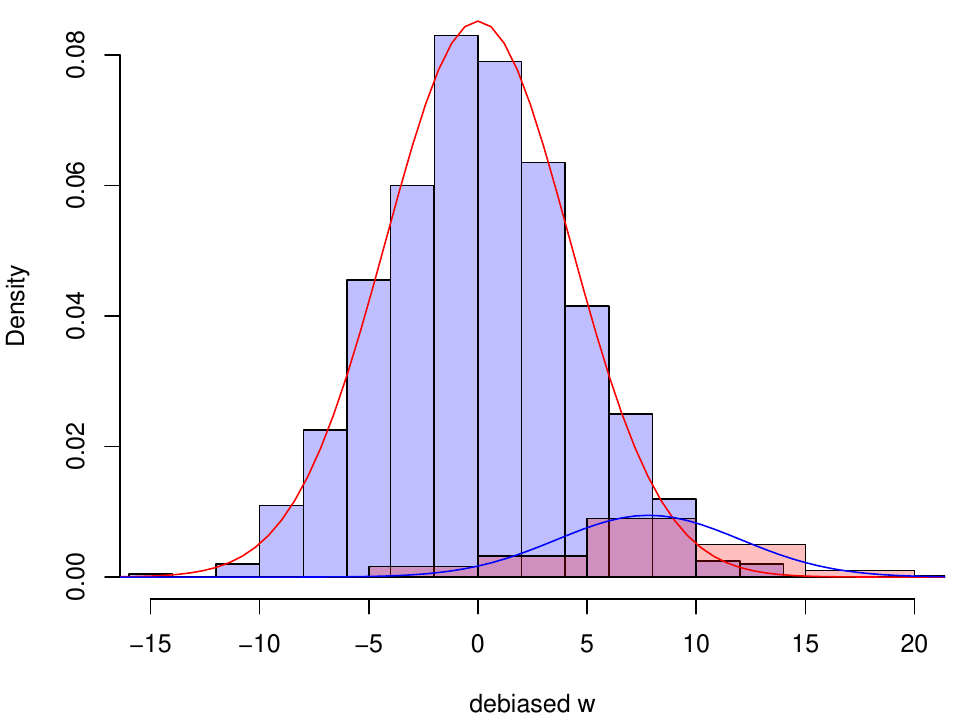}
\end{center}
\caption{Histograms of the components of PLR estimator $\hat\bw$ (left) and the corresponding de-biased estimator $\bar\bw$ (right) for a typical dataset. In the right plot, the curves represent the asymptotic normal densities for the zero and nonzero components.}
\label{fig:normal}
\end{figure}

This subsection evaluates the performance of variable selection through confidence intervals constructed using the formulas proposed in Section~\ref{section:method}. Synthetic data are generated following the same procedure as in Section~\ref{example:precision}. For a typical dataset, the histogram of the estimated coefficients $\hat \bw$ is displayed in the left plot of Figure~\ref{fig:normal}, which corresponds to the AR1 case with $\epsilon = 0.1$ and $\log\lambda = -2$. The spike indicates that most components of $\hat \bw$ are zero, which is an effect of the LASSO penalty. The right plot of Figure~\ref{fig:normal} shows the histogram $\bar \bw$, which is a de-biased estimate. The histogram shows a slightly right-skewed distribution resulting from two closely positioned clusters. Note that the mean of $\bar\bw$ is $\sqrt{p} R_0 \bSigma^{-1}\hat\mu/\zeta$, which corresponds to the first term in (\ref{df2}) and takes two distinct values.  
The first cluster corresponds to zero coefficients, and the second cluster corresponds to the non-zero coefficients. Both clusters follow a normal distribution, as suggested in Proposition~\ref{prop:main}.

\begin{figure}[hb]
\begin{center}
\includegraphics[width = 0.48\textwidth]{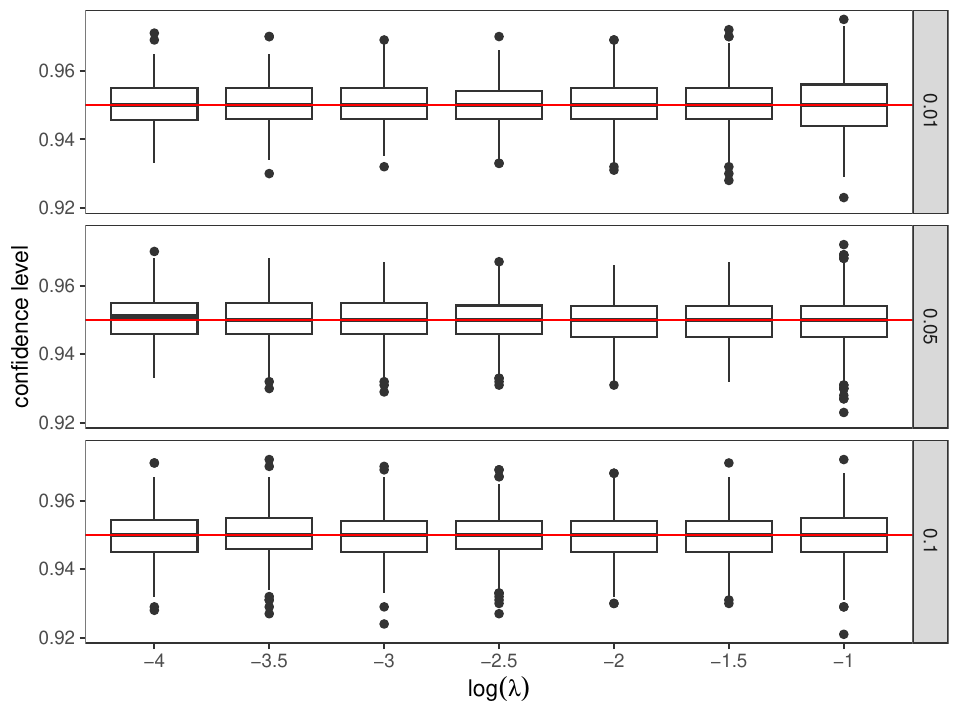}
\includegraphics[width = 0.48\textwidth]{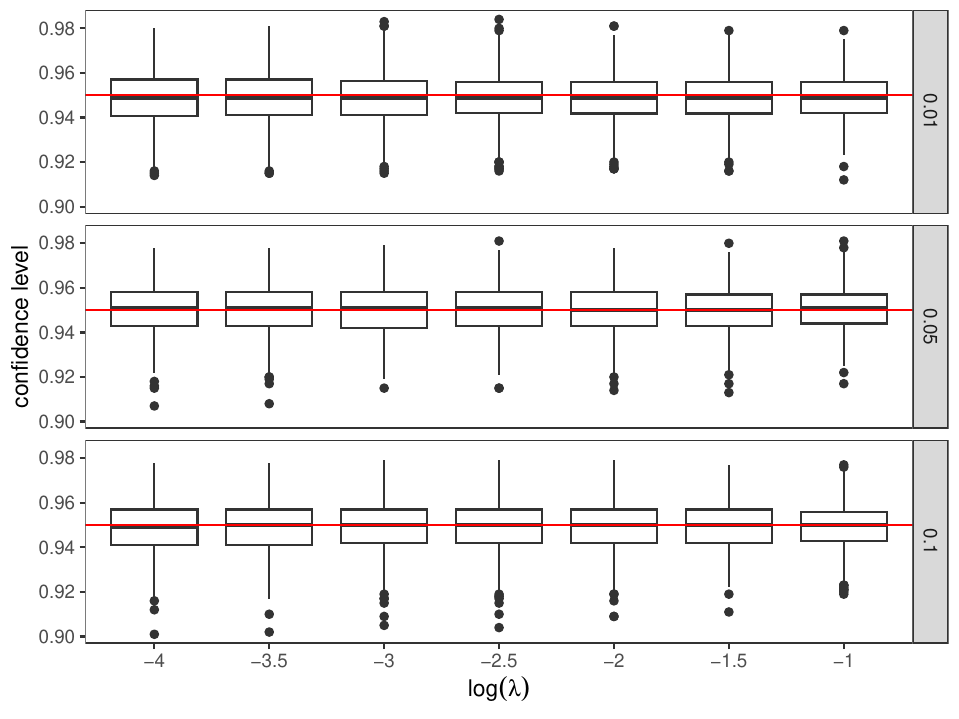}
\includegraphics[width = 0.48\textwidth]{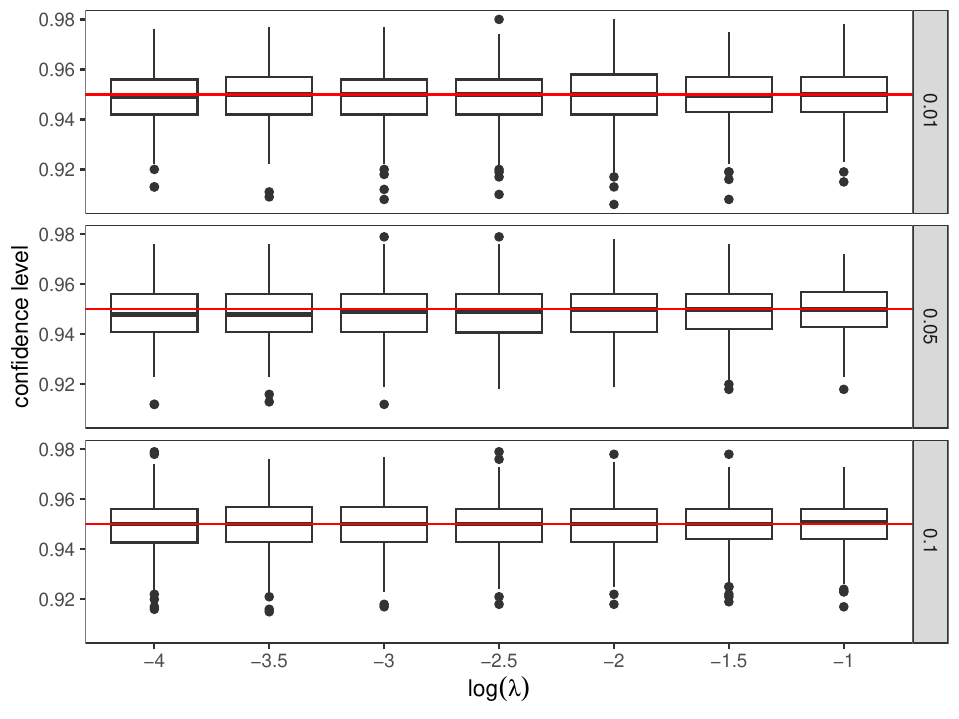}
\includegraphics[width = 0.48\textwidth]{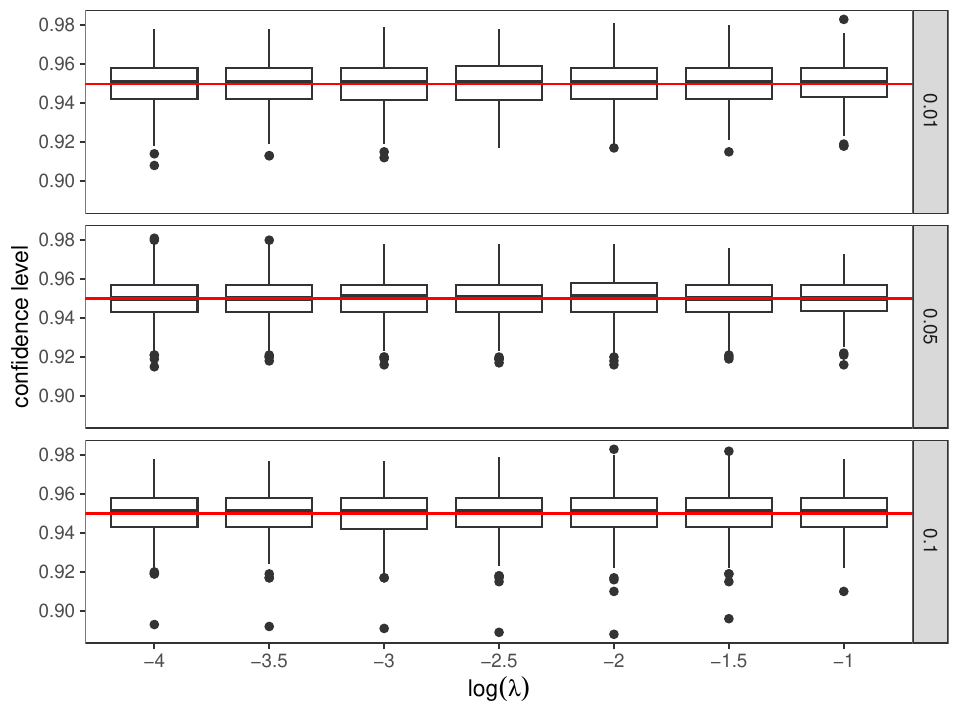}
\end{center}
\caption{Boxplots of empirical confidence levels based on 500 replicates for four correlation structures: IID (top-left), block (top-right), AR1 (bottom-left), and banded (bottom-right). In each plot, the horizontal line indicates the nominal confidence level. Different rows of plots represent different sparsity levels, $\epsilon = 0.01, 0.05, 0.1$.}
\label{fig:coverage}
\end{figure}

We further construct individual 95\% confidence intervals for components of $\bw$ following the formula (\ref{confi}) derived in Section~\ref{section:method}. Ideally, we expected 95\% of these confidence intervals to include the corresponding true parameters. Consider 4 correlation structures as explained in Section~\ref{example:precision}. Each box plot in Figure~\ref{fig:coverage} summarizes the empirical confidence levels based on 500 replicates for a scenario with a different sparsity level $\epsilon = 0.01, 0.05, 0.1$ or a tuning parameter $\log\lambda = -4, -3.5, \ldots, -1$. 
In these graphs, the horizontal line represents the nominal confidence level of 95\%. These findings confirm the validity of the formulas for constructing confidence intervals in various choices of $\lambda$. Although plots for other configurations demonstrate similar consistency, they are not displayed here to save space; nevertheless, they are available upon request.

\begin{figure}[hb]
\begin{center}
\includegraphics[width = 0.48\textwidth]{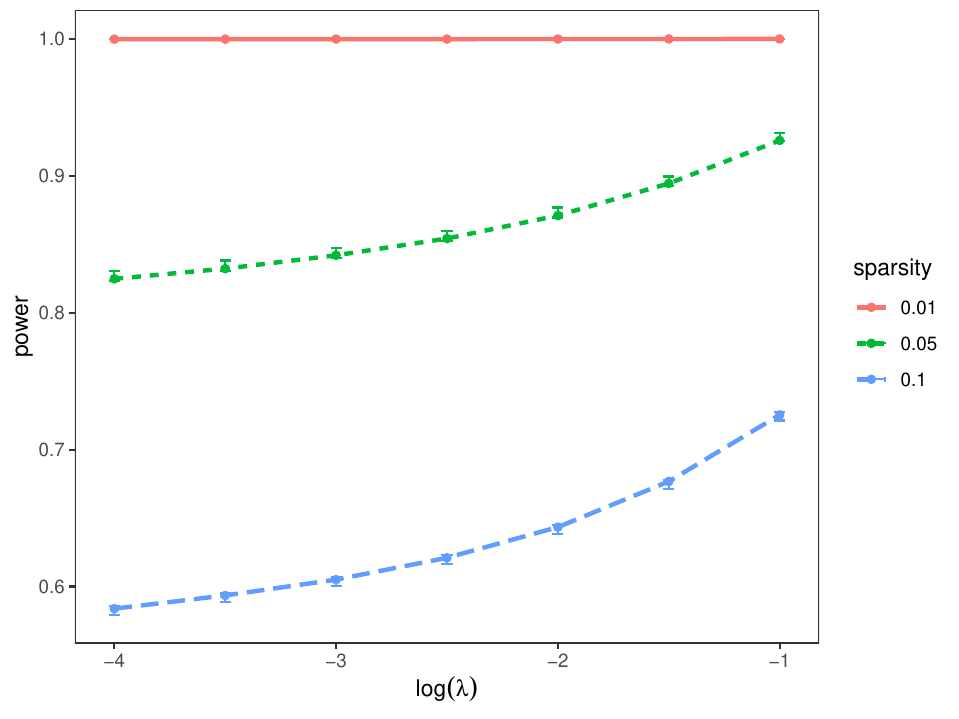}
\includegraphics[width = 0.48\textwidth]{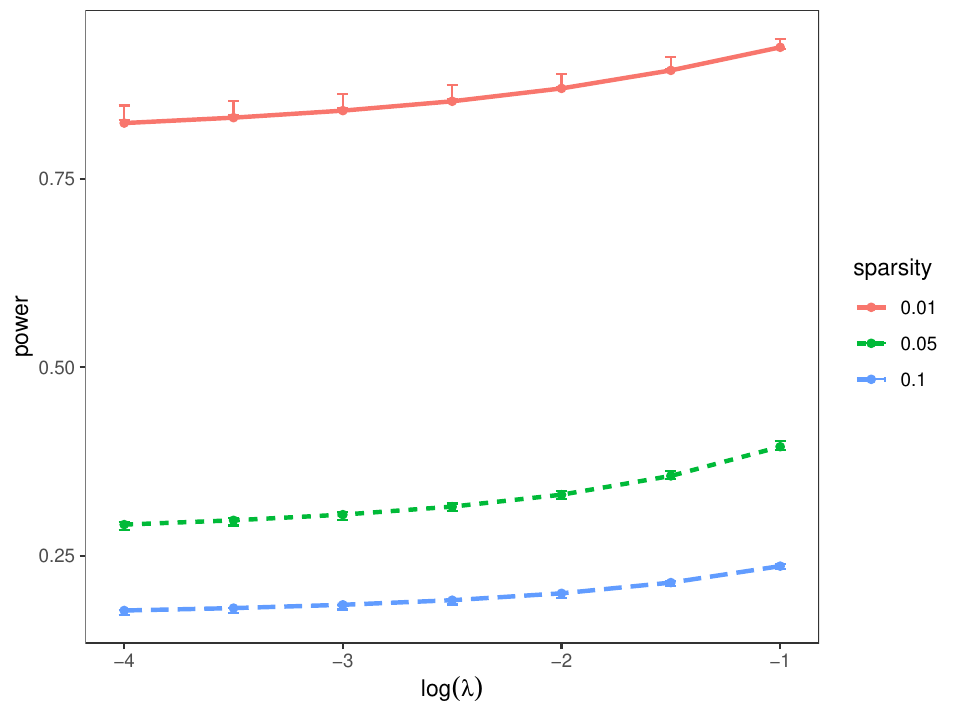}
\includegraphics[width = 0.48\textwidth]{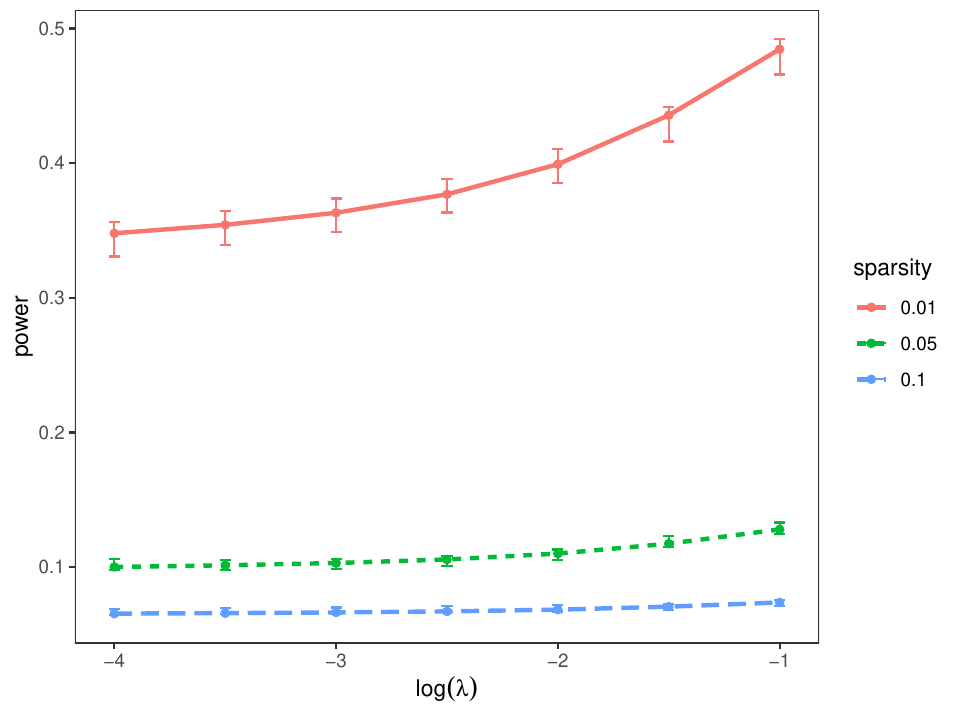}
\includegraphics[width = 0.48\textwidth]{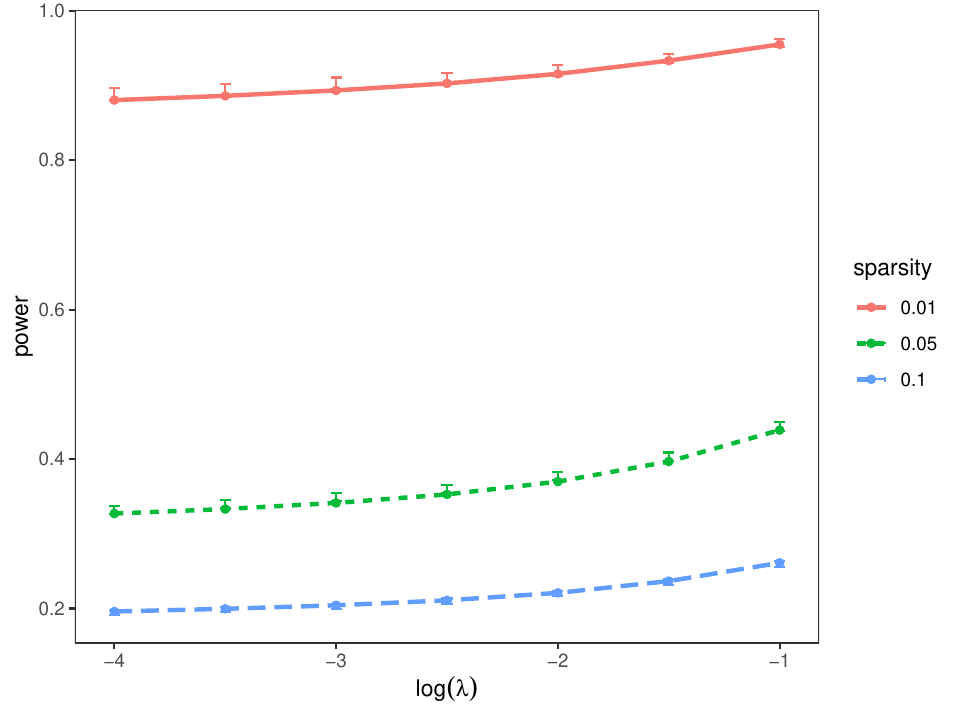}
\end{center}
\caption{Comparison between theoretical and empirical powers of hypothesis testing for four different correlation structures: IID (top-left), block (top-right), AR1 (bottom-left), and banded (bottom-right). In each plot, the three lines are the theoretical powers at different sparsity levels $\epsilon = 0.01, 0.05, 0.1$. The error bars are the 95\% confidence intervals of the mean powers based on 500 replicates.}
\label{fig:power}
\end{figure}

Furthermore, we assess the power of the procedure as the probability of correctly identifying significant variables when their corresponding true values are non-zero. The theoretical power is derived using Proposition~\ref{prop:main}, while the empirical power is estimated as the proportion of non-zero components whose 95\% confidence intervals do not include zero. Figure~\ref{fig:power} displays the theoretical powers as lines and 95\% confidence intervals of the mean empirical powers based on 500 replicates as error bars. The close alignment between the theoretical and empirical powers indicates the validity of the theory. The results show that the power increases with a larger value of $\lambda$, which demonstrates a pattern similar to the precision rate in Figure~\ref{fig:precision}.
Furthermore, the sparsity level of $\bw$ also influences the power; specifically, the power decreases as $\bw$ becomes denser. 

\section{Discussion}
\label{dis}

This paper focuses on learning Gaussian mixture data using $L_1$-regularized classification methods. We study the asymptotic behavior of the estimators in the framework in which $p,n\rightarrow\infty$ while $n/p\rightarrow\alpha$ is fixed. We first derive the limiting distribution of the regularized convex classifiers with an arbitrarily covariance structure. Then we obtain and the generalization error of the classifiers and further propose a de-biased estimator for classification, which allows us to perform variable selection through an appropriate hypothesis testing procedure. Using $L_1$-regularized logistic regression as an example, we conduct extensive computational experiments to confirm that our theoretical predictions are consistent with simulation results. Our next step is to implement the current framework for another commonly used classification method: the support vector machine. 

Our analysis is based on the replica method, which is not yet fully rigorous. Another future direction of our research is to provide a rigorous justification for the results derived in this paper. So far, the rigorous work in this area has mainly focused on i.i.d. randomness. The rigorous results for general covariance structure are more challenging, and one possible solution is to use Gordon’s Gaussian minmax inequalities \citep{Gordon1985}. 

\renewcommand{\theequation}{A-\arabic{equation}}
  \setcounter{equation}{0}
\section*{Appendix}

\subsection*{Derivation of Proposition \ref{prop:main}}
This appendix outlines the replica calculation that leads to Proposition 1. We limit ourselves to the main steps. For a general introduction to the replica method and its motivation, we refer to \cite{mezard1987spin,mezard2009information}. 

Denote $\bX=[\bx_1,\cdots,\bx_n]^T$, $\by=(y_1,\cdots,y_n)^T$. We consider regularized classification of the form
\begin{eqnarray}\label{classreg}
\hat{\bw}&=&\text{argmin}_{\bw}\left\{\sum_{i=1}^nV\left(\frac{y_i\bx_i^T\bw}{\sqrt{p}}\right)+\sum_{j=1}^pJ_\lambda(w_j)\right\}.
\end{eqnarray}
After suitable scaling, the terms inside the bracket $\{\cdot\}$ are exactly equal to the objective function of the model (\ref{class}) in the main text. 

The replica calculation aims at estimating the following moment generating function (partition function)
\begin{eqnarray}\nn
Z_p(\beta,s)&=&\int\exp\left\{-\beta\left[\sum_{i=1}^nV\left(\frac{y_i\bx_i^T(\bw+s\tilde{d}\bSigma^{-1}\bv)}{\sqrt{p}}\right)\right.\right.\\\label{partition}
&&\left.\left.+\sum_{j=1}^p\left\{J_\lambda(w_j)-s(g(v_j)-v_jw_j)\right\}\right]\right\}d\bw d\bv,
\end{eqnarray}
where $\tilde{d}\in\mR$ will be defined below, $\beta\textgreater 0$ is a `temperature' parameter, $s\textgreater 0$, and $g:\mR\rightarrow\mR$ a continuous function strictly convex in its first argument. In the zero temperature limit, i.e. $\beta\rightarrow\infty$, $Z_p(\beta,s=0)$ is dominated by the values of $\bw$ and $w_0$ which are the solution of (\ref{classreg}). 

Within the replica method, it is assumed that the limits $p\rightarrow\infty$, $\beta\rightarrow\infty$ exist almost surely for the quantity $(p\beta)^{-1}\log Z_p(\beta,s)$, and that the order of the limits can be exchanged. We therefore define the free energy. 
\begin{eqnarray}\label{calf1}
{\cal F}(s)&=&-\lim_{\beta\rightarrow\infty}\lim_{p\rightarrow\infty}\frac{1}{p\beta}\log Z_p(\beta,s)=-\lim_{p\rightarrow\infty}\lim_{\beta\rightarrow\infty}\frac{1}{p\beta}\log Z_p(\beta,s).
\end{eqnarray}
In other words, ${\cal F}(s)$ is the exponential growth rate of $Z_p(\beta,s)$. It is also assumed that $p^{-1}\log Z_p(\beta,s)$ concentrate tightly around its expectation so that ${\cal F}(s)$ can in fact be evaluated by computing 
\begin{eqnarray}\label{calf1}
{\cal F}(s)&=&-\lim_{p\rightarrow\infty}\lim_{\beta\rightarrow\infty}\frac{1}{p\beta}\bE\log Z_p(\beta,s),
\end{eqnarray}
where the expectation is with respect to the distribution of training data $\bX$ and $\by$. Notice that, by (\ref{calf1}) and using Laplace method in the integral (\ref{partition}), we have
\begin{eqnarray}\nn
{\cal F}(s)&=&\lim_{p\rightarrow\infty}\frac{1}{p}\min_{\bw,\bv}\left\{\sum_{i=1}^nV\left(\frac{y_i\bx_i^T(\bw+s\tilde{d}\bSigma^{-1}\bv)}{\sqrt{p}}\right)+\sum_{j=1}^p\left\{J_\lambda(w_j)+s[g(v_j)-v_jw_j]\right\}\right\}.
\end{eqnarray}
We assume that the derivative of ${\cal F}(s)$ as $s\rightarrow 0$ can be obtained by differentiating inside the limit. This condition holds, for example, if the cost function is strongly convex at $s=0$. We get
\begin{eqnarray}\nn
\frac{d{\cal F}}{ds}(s=0)&=&\lim_{p\rightarrow\infty}\frac{1}{p}\min_{\bw,\bv}\left\{\sum_{i=1}^nV^\prime\left(\frac{y_i\bx_i^T\bw}{\sqrt{p}}\right)\frac{\tilde{d}y_i\bx_i^T\bSigma^{-1}\bv}{\sqrt{p}}+\sum_{j=1}^p\left\{[g(v_j)-v_jw_j]\right\}\right\}\\\nn
&=&\lim_{p\rightarrow\infty}\frac{1}{p}\min_{\bw,\bv}\left\{\sum_{j=1}^p\left[g(v_j)-v_j\left\{w_j-\sum_{i=1}^nV^\prime\left(\frac{y_i\bx_i^T\bw}{\sqrt{p}}\right)\frac{\tilde{d}y_i(\bx_i^T\bSigma^{-1})_j}{\sqrt{p}}\right\}\right]\right\}\\\label{deriv}
&=&-\lim_{p\rightarrow\infty}\frac{1}{p}\sum_{j=1}^p\tilde{g}(\bar{w}_j),
\end{eqnarray}
where 
\begin{eqnarray}\label{wbar}
\bar{w}_j&=&\hat{w}_j-\sum_{i=1}^nV^\prime\left(\frac{y_i\bx_i^T\hat{\bw}}{\sqrt{p}}\right)\frac{\tilde{d}y_i(\bx_i^T\bSigma^{-1})_j}{\sqrt{p}}
\end{eqnarray}
for $j=1,\cdots,p$ and $\tilde{g}:\mR\times\mR\rightarrow\mR$ is the Lagrange dual of $g$ defined as $\tilde{g}(u)\equiv\max_{x\in\mR}[ux-g(x)]$. Hence, by computing ${\cal F}(s)$ using (\ref{calf1}) for a complete set of functions $\tilde{g}$, we get access to the corresponding limit quantities (\ref{deriv}), and hence, via standard weak convergence arguments, to the joint empirical distribution of $(\bar{w}_j,w_{0j})$.

In order to evaluate the integration of a log function, we make use of the replica method based on the identity 
\begin{eqnarray}
\log Z=\lim_{k\rightarrow 0}\frac{\partial Z^k}{\partial k}=\lim_{k\rightarrow 0}\frac{\partial}{\partial k}\log Z^k,
\end{eqnarray}
and rewrite (\ref{calf1}) as
\begin{eqnarray}\label{replica}
{\cal F}(s)=-\lim_{\beta\rightarrow\infty}\lim_{p\rightarrow\infty}\frac{1}{p\beta}\lim_{k\rightarrow 0}\frac{\partial}{\partial k}\Xi_k(\beta),
\end{eqnarray}
where 
\begin{eqnarray}\label{xi0}
\Xi_k(\beta)=\langle\{Z_\beta(\beta,s)\}^k\rangle_{\bX,\by}=\int\{Z_\beta(\beta,s)\}^k\prod_{i=1}^nP(\bx_i,y_i)d\bx_i dy_i.
\end{eqnarray}
Equation (\ref{replica}) can be derived by using the fact that $\lim_{k\rightarrow 0}\Xi_k(\beta)=1$ and exchanging the order of 
the averaging and the differentiation with respect to $k$. In the replica method, we will first evaluate $\Xi_k(\beta)$ for the integer $k$ and then apply it to the real $k$ and take the limit of $k\rightarrow 0$. 

For the integer $k$, to represent $\{Z_\beta(\bX,\by)\}^k$ in the integrand of (\ref{xi0}), we use the identity.
\begin{eqnarray}\nn
\left(\int f(x)\nu(dx)\right)^k=\int f(x_1)\cdots f(x_k)\nu(dx_1)\cdots\nu(dx_k),
\end{eqnarray}
where $\nu(dx)$ denotes the measure over $x\in\mR$. We obtain
\begin{eqnarray}\nn
\{Z_\beta(\beta,s)\}^k&=&\prod_{a=1}^k\left(\int\exp\left\{-\beta\left[\sum_{i=1}^nV\left(\frac{y_i\bx_i^T(\bw^a+s\tilde{d}\bSigma^{-1}\bv^a)}{\sqrt{p}}\right)\right.\right.\right.\\\nn
&&\left.\left.\left.+\sum_{j=1}^p\left\{J_\lambda(w^a_j)+s(g(v^a_j,w_{0j})-v^a_jw^a_j)\right\}\right]\right\}d\bw^a d\bv^a\right),
\end{eqnarray}
where we have introduced replicated parameters
\begin{eqnarray}\nn
\bw^a\equiv[w^a_1,\cdots,w^a_p]^T&\text{and}&\bv^a\equiv[v^a_1,\cdots,v^a_p]^T, \text{ for }a=1,\cdots,k.
\end{eqnarray}
Exchanging the order of the two limits $p\rightarrow\infty$ and $k\rightarrow 0$ in (\ref{replica}), we have 
\begin{eqnarray}\label{freef}
{\cal F}(s)=-\lim_{\beta\rightarrow\infty}\frac{1}{\beta}\lim_{k\rightarrow 0}\frac{\partial}{\partial k}\left(\lim_{p\rightarrow\infty}\frac{1}{p}\Xi_k(\beta)\right).
\end{eqnarray}
Define the measure $\nu(d\bw)$ over $\bw\in\mathbb{R}^p$ as follows
\begin{eqnarray}\nn
\nu(d\bw)&=&\left[\int\exp\left\{-\beta\left[J_\lambda(\bw-s\tilde{d}\bSigma^{-1}\bv)+s\sum_{j=1}^p\left\{g(v_j)-v_jw_j+s\tilde{d}v_j(\bSigma^{-1}\bv)_j\right\}\right]\right\}d\bv\right]d\bw.
\end{eqnarray}
Similarly, define the measure $\nu_+(d\bx)$ and $\nu_-(d\bx)$ over $\bx\in\mathbb{R}^p$ as
\begin{eqnarray}\nn
\nu_+(d\bx)=P(\bx|y=+1)d\bx&\text{ and }&\nu_-(d\bx)=P(\bx|y=-1)d\bx. 
\end{eqnarray}
In order to carry out the calculation of $\Xi_k(\beta)$, we let $\nu^k(d\bw)\equiv\nu(d\bw^1)\times\cdots\times\nu(d\bw^k)$ be a measure over $(\mathbb{R}^p)^k$, with $\bw^1,\cdots,\bw^k\in\mathbb{R}^p$. Analogously $\nu^n(d\bx)\equiv\nu(d\bx_1)\times\cdots\times\nu(d\bx_n)$ with $\bx_1,\cdots,\bx_n\in\mathbb{R}^p$, $\nu^n(dy)\equiv\nu(dy_1)\times\cdots\times\nu(dy_n)$ with $y_1,\cdots,y_n\in\{-1,1\}$. With these notations and the change of variable $\bw^a+s\tilde{d}\bSigma^{-1}\bv^a\rightarrow\bw^a$, we can rewrite (\ref{xi0}) as
\begin{eqnarray}\nn
\Xi_k(\beta)&=&\int\exp\left\{-\beta\sum_{i=1}^n\sum_{a=1}^kV\left(\frac{y_i\bx_i^T\bw^a}{\sqrt{p}}\right)\right\}\nu^k(d\bw)\nu^n(d\by)\nu^n(d\bx)\\\nn
&=&\int\left[\int\exp\left\{-\beta\sum_{a=1}^kV\left(\frac{\bx^T\bw^a}{\sqrt{p}}\right)\right\}\nu_+(d\bx)\right]^{n_+}\times\\\nn
&&~~~\left[\int\exp\left\{-\beta\sum_{a=1}^kV\left(\frac{-\bx^T\bw^a}{\sqrt{p}}\right)\right\}\nu_-(d\bx)\right]^{n_-}\nu^k(d\bw)\\\label{xi}
&=&\int\exp\{p(\alpha_+\log I_++\alpha_-\log I_-)\}\nu^k(d\bw),
\end{eqnarray}
where $\alpha_\pm=n_\pm/p$ and
\begin{eqnarray}\label{iterm}
I_\pm&=&\int\exp\left\{-\beta\sum_{a=1}^kV\left(\frac{\pm\bx^T\bw^a}{\sqrt{p}}\right)\right\}\nu_\pm(d\bx).
\end{eqnarray}
Notice that above we used the fact that the integral over $(\bx_1,\cdots,\bx_n)\in(\mathbb{R}^p)^n$ factors into $n_+$ integrals over $(\mathbb{R})^p$ with measure $\nu_+(d\bx)$ and $n_-$ integrals over $(\mathbb{R})^p$ with measure $\nu_-(d\bx)$. Next, we introduce the integration variables $du^a,d\tilde{u}^a$ for $1\le a\le k$. Letting $\nu^k(du)=du^1\cdots du^k$ and $\nu^k(d\tilde{u})=d\tilde{u}^1\cdots d\tilde{u}^k$, we obtain
\begin{eqnarray}\nn
I_\pm&=&\int\exp\left\{-\beta\sum_{a=1}^kV\left(u^a\right)\right\}\prod_{a=1}^k\delta\left(u^a\mp\frac{\bx^T\bw^a}{\sqrt{p}}\right)\nu^k(du)\nu_\pm(d\bx)\\\nn
&=&\int\exp\left\{-\beta\sum_{a=1}^kV\left(u^a\right)\right\}\exp\left\{\sum_{a=1}^ki\sqrt{p}\left(u^a\mp\frac{\bx^T\bw^a}{\sqrt{p}}\right)\tilde{u}^a\right\}\nu^k(du)\nu^k(d\tilde{u})\nu_\pm(d\bx)\\\nn
&=&\int\exp\left\{-\beta\sum_{a=1}^kV\left(u^a\right)+i\sqrt{p}\sum_{a=1}^k\left(u^a\mp\frac{\bx^T\bw^a}{\sqrt{p}}\right)\tilde{u}^a\right\}\nu_\pm(d\bx)\nu^k(du)\nu^k(d\tilde{u})\\\nn
&=&\int\exp\left\{-\beta\sum_{a=1}^kV\left(u^a\right)+i\sqrt{p}\sum_{a=1}^ku^a\tilde{u}^a-\frac{1}{2}\sum_{ab}(\bw^a)^T\bSigma\bw^b\tilde{u}^a\tilde{u}^b\right.\\\label{iplus}
&&~~~~~~~~~~~~~~~~~~~~~~~~~~~\left.-i\sum_{a=1}^k(\bw^a)^T\bmu\tilde{u}^a\right\}\nu^k(du)\nu^k(d\tilde{u}).
\end{eqnarray}
Note that conditional on $y=\pm 1$, $\bx$ follows multivariate distributions with mean $\pm\bmu$ and covariance matrix $\bSigma$. In deriving (\ref{iplus}), we have used the fact that the low-dimensional marginals of $\bx$ can be approximated by a Gaussian distribution based on the multivariate central limit theorem.  

From (\ref{iplus}) we have $I_+=I_-\equiv I$. Thus, applying (\ref{iplus}) to (\ref{xi}) we obtain
\begin{eqnarray}\nn
\Xi_k(\beta)&=&\int\exp\{p\alpha\log I\}\nu^k(d\bw).
\end{eqnarray}
Now we introduce integration variables $Q_{ab},\tilde{Q}_{ab}$ and $R^a,R_0^a$ associated with $(\bw^a)^T\bSigma\bw^b/p$ and $(\bw^a)^T\hat{\bmu}/\sqrt{p}$ respectively for $1\le a,b\le k$ where $\hat{\bmu}=\bmu/\mu$. Denote $\bQ\equiv(Q_{ab})_{1\le a,b\le k}$, $\tilde{\bQ}\equiv(\tilde{Q}_{ab})_{1\le a,b\le k}$, $\bR\equiv(R^a)_{1\le a\le k}$, and $\bR_0\equiv(R_0^a)_{1\le a\le k}$. Note that constant factors can be applied to the integration variables, and we choose convenient factors for later calculations.  Letting $d\bQ\equiv\prod_{a,b}dQ_{ab}$, $d\tilde{\bQ}\equiv\prod_{a,b}d\tilde{Q}_{ab}$, $d\bR\equiv\prod_{a}dR^{a}$, and $d\bR_0\equiv\prod_{a}dR_0^{a}$, we obtain
\begin{eqnarray}\nn
\Xi_k(\beta)&=&\int\exp\{p\alpha\log I\}\nu^k(d\bw)\\\nn
&&\int\exp\left\{i\beta p\left(\sum_{ab}\left[Q_{ab}-\frac{(\bw^a)^T\bSigma\bw^b}{p}\right]\tilde{Q}_{ab}+\sum_a\left[R^{a}-\frac{(\bw^a)^T\hat{\bmu}}{\sqrt{p}}\right]R_0^{a}\right)\right\}d\bQ d\tilde{\bQ}d\bR d\bR_0\\\nn
&=&\int\exp\left\{-p\left[-\alpha\log I-i\beta\left\{\sum_{ab}\left[Q_{ab}-\frac{(\bw^a)^T\bSigma\bw^b}{p}\right]\tilde{Q}_{ab}+\sum_a\left[R^{a}-\frac{(\bw^a)^T\hat{\bmu}}{\sqrt{p}}\right]R_0^{a}\right\}\right]\right\}\\\nn
&&\nu^k(d\bw)
d\bQ d\tilde{\bQ}d\bR d\bR_0\\\nn
&=&\int\exp\left\{-p\left[-\log I-i\beta\left(\sum_{ab}Q_{ab}\tilde{Q}_{ab}+\sum_aR^{a}R_0^{a}\right)\right]\right\}\\\nn
&&\left[\int\exp\left\{-i\beta\sum_{ab}\tilde{Q}_{ab}(\bw^a)^T\bSigma\bw^b-i\beta\sqrt{p}\sum_{a}R_0^{a}(\bw^a)^T\hat{\bmu}\right\}\nu^k(d\bw)\right]
d\bQ d\tilde{\bQ}d\bR d\bR_0,
\end{eqnarray}
which can be rewritten as
\begin{eqnarray}\label{saddle}
\Xi_k(\beta)&=&\int\exp\left\{-p{\cal S}_k(\bQ,\tilde{\bQ},\bR,\bR_0)\right\}d\bQ d\tilde{\bQ}d\bR d\bR_0,
\end{eqnarray}
where
\begin{eqnarray}\nn
{\cal S}_k(\bQ,\tilde{\bQ},\bR,\bR_0)&=&-i\beta\left(\sum_{ab}Q_{ab}\tilde{Q}_{ab}+\sum_aR^{a}R_0^{a}\right)-\frac{1}{p}\log\xi(\tilde{\bQ},\bR_0)-\hat{\xi}(\bQ,\bR),
\end{eqnarray}
where
\begin{eqnarray}\nn
\xi(\tilde{\bQ},\bR_0)&=&\int\exp\left\{-i\beta\sum_{ab}\tilde{Q}_{ab}(\bw^a)^T\bSigma\bw^b-i\beta\sum_a\sqrt{p}R_0^a(\bw^a)^T\hat{\bmu}\right\}\nu^k(d\bw),\\\label{xihat}
\hat{\xi}(\bQ,\bR)&=&\alpha\log\hat{I},
\end{eqnarray}
where $\hat{I}$ can be obtained from (\ref{iplus}) as
\begin{eqnarray}\nn
\hat{I}&=&\int\exp\left\{-\beta\sum_{a=1}^kV\left(u^a\right)+i\sqrt{p}\sum_{a=1}^ku^a\tilde{u}^a\right.\\\label{ihat}
&&~~~~~~~~~~\left.-\frac{p}{2}\sum_{ab}Q_{ab}\tilde{u}^a\tilde{u}^b-i\sqrt{p}\sum_{a=1}^kR^a\mu\tilde{u}^a\right\}\nu^k(du)\nu^k(d\tilde{u}).
\end{eqnarray}
Now we apply steepest descent method to the remaining integration. According to Varadhan's proposition \citep{Tanaka}, only the saddle points of the exponent of the integrand contribute to the integration in the limit of $p\rightarrow\infty$. We next use the saddle point method in (\ref{saddle}) to obtain
\begin{eqnarray}\nn
-\lim_{p\rightarrow\infty}\frac{1}{p}\Xi_k(\beta)&=&{\cal S}_k(\bQ^\star,\tilde{\bQ}^\star,\bR^\star,\bR_0^\star),
\end{eqnarray}
where $\bQ^\star,\tilde{\bQ}^\star,\bR^\star,\bR_0^\star$ are the saddle point locations. Looking for saddle-points over all the entire space is in general difficult to perform. We  assume replica symmetry for saddle-points such that they are invariant under exchange of any two replica indices $a$ and $b$, where $a\ne b$. Under this symmetry assumption, the space is greatly reduced and the exponent of the integrand can be explicitly evaluated.  The replica symmetry is also motivated by the fact that ${\cal S}_k(\bQ^\star,\tilde{\bQ}^\star,\bR^\star,\bR_0^\star)$ is indeed left unchanged by such change of variables. This is equivalent to postulating that $R^a=R$, $R_0^a=iR_0$, 
\begin{eqnarray}\label{qsym}
(Q_{ab})^\star=\left\{\begin{array}{cc}q_1&\text{if a=b}\\q_0&\text{otherwise}\end{array}\right.,&\text{and}&(\tilde{Q}_{ab})^\star=\left\{\begin{array}{cc}i\frac{\beta\zeta_1}{2}&\text{if a=b}\\i\frac{\beta\zeta_0}{2}&\text{otherwise}\end{array}\right.,
\end{eqnarray}
where the factor $i\beta/2$ is for future convenience. The next step consists of substituting the above expressions for $\bQ^\star,\tilde{\bQ}^\star,\bR^\star,\bR_0^\star$ in ${\cal S}_k(\bQ^\star,\tilde{\bQ}^\star,\bR^\star,\bR_0^\star)$ and then taking the limit $k\rightarrow 0$. We will separately consider each term of ${\cal S}_k(\bQ^\star,\tilde{\bQ}^\star,\mR^\star,R_0^\star)$. Let us begin with the first term.
\begin{eqnarray}\label{qqhat}
-i\beta\left(\sum_{ab}Q_{ab}\tilde{Q}_{ab}+\sum_aR^aR_0^{a}\right)&=&\frac{k\beta^2}{2}(\zeta_1q_1-\zeta_0q_0)+k\beta RR_0+o(k).
\end{eqnarray}
Next consider $\log\xi(\tilde{\bQ},\bR_0)$. For p-vectors $\bu,\bv\in\mR^p$ and $p\times p$ matrix $\bSigma$, introducing the notation $\|\bv\|_{\bSigma}^2\equiv \bv^T\bSigma\bv$ and $\langle\bu,\bv\rangle\equiv\sum_{j=1}^pu_jv_j/p$, we have
\begin{eqnarray}\nn
\xi(\tilde{\bQ},\bR_0)&=&\int\exp\left\{\frac{\beta^2}{2}(\zeta_1-\zeta_0)\sum_{a=1}^k\|\bw^a\|^2_{\bSigma}+\frac{\beta^2\zeta_0}{2}\sum_{a,b=1}^k(\bw^a)^T\bSigma\bw^b\right.\\\nn
&&~~~~~~~~~\left.+\beta\sqrt{p}\sum_{a=1}^kR_0(\bw^a)^T\hat{\bmu}\right\}\nu^k(d\bw)\\\nn
&=&E\int\exp\left\{\frac{\beta^2}{2}(\zeta_1-\zeta_0)\sum_{a=1}^k\|\bw^a\|^2_{\bSigma}+\beta\sqrt{\zeta_0}\sum_{a=1}^k(\bw^a)^T\bSigma^{1/2}\bz\right.\\\label{xiq}
&&~~~~~~~~~\left.+\beta\sqrt{p}\sum_{a=1}^kR_0(\bw^a)^T\hat{\bmu}\right\}\nu^k(d\bw),
\end{eqnarray}
where expectation is with respect to $\bz\sim N(0,\bI_{p\times p})$. Notice that, given $\bz\in \mR^p$, the integrals over $\bw^1,\cdots,\bw^k$ factorize, whence
\begin{eqnarray}\nn
\xi(\tilde{\bQ},\bR_0)&=&E\left\{\left[\int\exp\left\{\frac{\beta^2}{2}(\zeta_1-\zeta_0)\|\bw\|^2_{\bSigma}+\beta\sqrt{\zeta_0}\bw^T\bSigma^{1/2}\bz\right.\right.\right.\\\nn
&&\left.\left.\left.+\beta\sqrt{p}R_0\bw^T\hat{\bmu}\right\}\nu(d\bw)\right]^k\right\}.
\end{eqnarray}
Finally, after integration over $\nu^k(d\tilde{u})$, (\ref{ihat}) becomes
\begin{eqnarray}\nn
\hat{I}&=&\int\exp\left\{-\beta\sum_{a=1}^kV\left(u^a\right)-\frac{1}{2}\sum_{ab}(u^a-R\mu)(\bQ^{-1})_{ab}(u^b-R\mu)\right.\\\label{ihat1}
&&~~~~~~~~~\left.-\frac{1}{2}\log\text{det}\bQ\right\}\nu^k(du).
\end{eqnarray}
Similarly, using (\ref{qsym}), we obtain
\begin{eqnarray}\nn
\sum_{ab}(u^a-R\mu)(\bQ^{-1})_{ab}(u^b-R\mu)&=&\frac{\beta\sum_a(u^a-R\mu)^2}{q}-\frac{\beta^2q_0\{\sum_a(u^a-R\mu)\}^2}{q^2},\\\nn
\log\text{det}\bQ&=&\log\left[(q_1-q_0)^k\left(1+\frac{kq_0}{q_1-q_0}\right)\right]=\frac{k\beta q_0}{q},
\end{eqnarray}
where we retain only the leading order terms. Therefore, (\ref{ihat1}) becomes
\begin{eqnarray}\nn
\hat{I}&=&\int\exp\left\{-\beta\sum_{a=1}^kV\left(u^a\right)-\frac{\beta\sum_a(u^a-R\mu)^2}{2q}+\frac{\beta^2q_0\{\sum_a(u^a-R\mu)\}^2}{2q^2}\right.\\\nn
&&~~~~~~~~~\left.-\frac{1}{2}\frac{k\beta q_0}{q}\right\}\nu^k(du)\\\nn
&=&\int Dz\int\exp\left\{-\beta\sum_{a=1}^kV\left(u^a\right)-\frac{\beta\sum_a(u^a-R\mu)^2}{2q}+\frac{\beta\sum_a(u^a-R\mu)\sqrt{q_0}z}{q}\right.\\\nn
&&~~~~~~~~~\left.-\frac{1}{2}\frac{k\beta q_0}{q}\right\}\nu^k(du)\\\nn
&=&\exp\left(-\frac{k\beta q_0}{2q}\right)\int Dz\left(\int\exp\left\{-\beta V(u)-\frac{\beta(u-R\mu-\sqrt{q_{0}}z)^2}{2q}+\frac{\beta q_{0}z^2}{2q}\right\}du\right)^k,
\end{eqnarray}
where the expectation $Dz=\int\frac{dz}{\sqrt{2\pi}}\exp\left(-\frac{z^2}{2}\right)$. Substituting this expression into (\ref{xihat}), we obtain
\begin{eqnarray}\nn
\hat{\xi}(\bQ,\bR)&=&\alpha\log\hat{I}=-\frac{k\alpha\beta q_0}{2q}+\alpha\log\int Dz\exp(k\log G)\\\nn
&=&-\frac{k\beta\alpha q_0}{2q}+\alpha\log\left(1+k\int Dz\log G\right)\\\label{ihat}
&=&-\frac{k\beta\alpha q_0}{2q}+k\alpha\int Dz\log G,
\end{eqnarray}
where 
\begin{eqnarray}\nn
\log G&=&\log\int\exp\left\{-\beta V(u)-\frac{\beta(u-R\mu-\sqrt{q_{0}}z)^2}{2q}+\frac{\beta q_{0}z^2}{2q}\right\}du\\\nn
&=&-\beta\min_u\left[V(u)+\frac{(u-R\mu-\sqrt{q_{0}}z)^2}{2q}-\frac{q_{0}z^2}{2q}\right]
\end{eqnarray}
in the limit of $\beta\rightarrow\infty$. Substituting this expression into (\ref{ihat}), we obtain 
\begin{eqnarray}\label{third}
\hat{\xi}(\bQ,\bR)&=&-k\alpha\beta E\left\{\min_u\left[V(u)+\frac{(u-R\mu-\sqrt{q_{0}}z)^2}{2q}\right]\right\},
\end{eqnarray}
where the expectation is with respect to $z\sim N(0,1)$. 

Taking the limit $\beta\rightarrow\infty$, the analysis of the saddle point parameters $q_0,q_1,\zeta_0,\zeta_1$ shows that $q_0,q_1$ has the same limit as $q_1-q_0=(q/\beta)+o(\beta^{-1})$ and $\zeta_0,\zeta_1$ has the same limit as $\zeta_1-\zeta_0=(-\zeta/\beta)+o(\beta^{-1})$. 
\begin{eqnarray}\nn
-i\beta\left(\sum_{ab}Q_{ab}\tilde{Q}_{ab}+\sum_aR^aR_0^{a}\right)&=&\frac{k\beta^2}{2}((\zeta_0-\zeta/\beta)(q_0+q/\beta)-\zeta_0q_0)+k\beta RR_0\\\label{qqhat1}
&=&\frac{k\beta}{2}(\zeta_0q-\zeta q_0)+k\beta RR_0.
\end{eqnarray}
\begin{eqnarray}\nn
\log\xi(\tilde{\bQ},\bR_0)&=&\log E\left\{\left[\int\exp\left\{-\beta\frac{\zeta}{2}\|\bw\|^2_{\bSigma}+\beta\sqrt{\zeta_0}\bw^T\bSigma^{1/2}\bz+\beta\sqrt{p}R_0\bw^T\hat{\bmu}\right\}\nu(d\bw)\right]^k\right\}\\\nn
&=&\text{E}\min_{\bw,\bv\in \mR^p}\left\{\frac{\zeta}{2}\|\bw\|_{\bSigma}^2-\left\langle\sqrt{\zeta_0}\bSigma^{1/2}\bz+\sqrt{p}R_0\hat{\bmu},\bw\right\rangle\right.\\\nn
&&\left.+\sum_{j=1}^p\left[J_\lambda(w_j-s\tilde{d}(\bSigma^{-1}\bv)_j)+s\left\{g(v_j)-v_jw_j+s\tilde{d}v_j(\bSigma^{-1}\bv)_j\right\}\right]\right\}
\end{eqnarray}
After changing the variable $\bw-s\tilde{d}\bSigma^{-1}\bv\rightarrow\bw$, we obtain
\begin{eqnarray}\nn
\log\xi(\tilde{\bQ},\bR_0)
&=&-k\beta\text{E}\min_{\bw,\bv\in \mR^p}\left\{\frac{\zeta}{2}\|\bw+s\tilde{d}\bSigma^{-1}\bv\|_{\bSigma}^2-\left\langle\sqrt{\zeta_0}\bSigma^{1/2}\bz+\sqrt{p}R_0\hat{\bmu},\bw+s\tilde{d}\bSigma^{-1}\bv\right\rangle\right.\\\label{xiq1}
&&\left.+\sum_{j=1}^p\left[J_\lambda(w_j)+s\left\{g(v_j)-v_jw_j\right\}\right]\right\}
\end{eqnarray}

Putting (\ref{qqhat1}), (\ref{xiq1}), and (\ref{third}) together into (\ref{saddle}) and then into (\ref{replica}), we obtain 
\begin{eqnarray}\nn
{\cal F}(s)&=&\frac{\zeta_0q-\zeta q_0}{2}+RR_0+\alpha\text{E}\min_{u\in\mR}\left\{V(u)+\frac{\left(u-R\mu-\sqrt{q_0}z\right)^2}{2q}\right\}\\\nn
&&+\frac{1}{p}\text{E}\min_{\bw,\bv\in \mR^p}\left\{\frac{\zeta}{2}\|\bw+s\tilde{d}\bSigma^{-1}\bv\|_{\bSigma}^2-\left\langle\sqrt{\zeta_0}\bSigma^{1/2}\bz+\sqrt{p}R_0\hat{\bmu},\bw+s\tilde{d}\bSigma^{-1}\bv\right\rangle\right.\\\label{fs}
&&\left.+\sum_{j=1}^p\left[J_\lambda(w_j)+s\left\{g(v_j)-v_jw_j\right\}\right]\right\},
\end{eqnarray}
where the expectations are with respect to $ z\sim N(0,1)$, and $\bz\sim N(0,\bI_{p\times p})$, with $z$ and $\bz$ independent from each other. 
\begin{eqnarray}\nn
&&\frac{d{\cal F}}{ds}(s=0)\\\nn
&=&\lim_{p\rightarrow\infty}\frac{1}{p}\min_{\bv\in\mR^p}E\left\{\zeta\tilde{d}\hat{\bw}^T\bv-\left\langle\sqrt{\zeta_0}\bSigma^{1/2}\bz+\sqrt{p}R_0\hat{\bmu},\tilde{d}\bSigma^{-1}\bv\right\rangle+\sum_{j=1}^p\left[g(v_j)-v_j\hat{w}_j\right]\right\},
\end{eqnarray}
where
\begin{eqnarray}\label{what}
\hat{\bw}&=&\text{argmin}_{\bw\in\mR^p}\left\{\frac{\zeta}{2}\left\|\bw-\frac{\sqrt{\zeta_0}}{\zeta}\bSigma^{-1/2}\bz-\frac{\sqrt{p}R_0\bSigma^{-1}\hat{\bmu}}{\zeta}\right\|_{\bSigma}^2+\sum_{j=1}^pJ_\lambda(w_j)\right\}.
\end{eqnarray}
At this point we choose $\tilde{d}=1/\zeta$. Minimizing over $\bv$ (recall that $\tilde{g}(x)=\max_{u\in\mR}[ux-g(u)]$, we get
\begin{eqnarray}\nn
&&\frac{d{\cal F}}{ds}(s=0)\\\nn
&=&\lim_{p\rightarrow\infty}\frac{1}{p}\min_{\bv\in\mR^p}E\left\{\sum_{j=1}^p\left[g(v_j,w_{0j})-v_j(\sqrt{\zeta_0}\bSigma^{-1/2}\bz+\sqrt{p}R_0\bSigma^{-1}\hat{\bmu})_j/\zeta\right]\right\}\\\nn
&=&-\lim_{p\rightarrow\infty}\frac{1}{p}E\left\{\sum_{j=1}^p\tilde{g}\left((\sqrt{\zeta_0}\bSigma^{-1/2}\bz+\sqrt{p}R_0\bSigma^{-1}\hat{\bmu})_j/\zeta\right)\right\}.
\end{eqnarray}
Renaming $\zeta_0=\zeta^2\tau^2$, we get out final expression for $\frac{d{\cal F}}{ds}(s=0)$ as
\begin{eqnarray}\label{df}
\frac{d{\cal F}}{ds}(s=0)&=&-\lim_{p\rightarrow\infty}\frac{1}{p}E\left\{\sum_{j=1}^p\tilde{g}\left((\tau\bSigma^{-1/2}\bz)_j+\sqrt{p}R_0(\bSigma^{-1}\hat{\bmu})_j/\zeta\right)\right\}.
\end{eqnarray}
Compared with (\ref{deriv}), this shows that the distribution limit of $\bar{\bw}$, defined in (\ref{wbar}), is the same as $\tau\bSigma^{-1/2}\bz+\sqrt{p}R_0\bSigma^{-1}\hat{\bmu}/\zeta$.

Here, $\zeta,\zeta_0,q,q_0,R,R_0$ are the order parameters which can be determined from the saddle-point equations of ${\cal F}(s=0)$. Define the functions $F$, $G$, and $H$ as 
\begin{eqnarray}\nn
F&=&\bE_z\left(\hat{u}-R\mu-\sqrt{q_0}z\right),\\\nn
G&=&\bE_z\left\{\left(\hat{u}-R\mu-\sqrt{q_0}z\right)z\right\},\\\nn
H&=&\bE_z\left\{\left(\hat{u}-R\mu-\sqrt{q_0}z\right)^2\right\},
\end{eqnarray}
where  
\begin{eqnarray}\nn
\hat{u}&=&\text{argmin}_{u\in\mR}\left\{V(u)+\frac{(u-R\mu-\sqrt{q_0}z)^2}{2q}\right\}.
\end{eqnarray}
Then all the order parameters can be determined by the following saddle-point equations derived from (\ref{fs}):
\begin{eqnarray}\label{eq01}
\xi_0&=&\frac{\alpha}{q^2}H,\\\label{eq02}
\xi&=&\frac{\alpha G}{\sqrt{q_0}q},\\\label{eq03}
q_0&=&\frac{1}{p}\bE_z\|\hat{\bw}\|^2_{\bSigma},\\\label{eq04}
q&=&\frac{1}{p\sqrt{\zeta_0}}\bE\left\langle\bSigma^{1/2}\bz,\hat{\bw}\right\rangle\\\label{eq05}
R&=&\frac{1}{\sqrt{p}}\bE_z\langle\hat{\bmu},\hat{\bw}\rangle,\\\label{eq06}
R_0&=&\frac{\alpha\mu}{q},
\end{eqnarray}
where $\hat{\bw}$ is solved from (\ref{what}). The result in (\ref{fs}) is for the general penalty function $J_\lambda(w)$. For quadratic penalty $J_\lambda(w)=\lambda w^2$, we get the closed form limiting distribution of $\hat{\bw}$ as 
\begin{eqnarray}\label{wlimit}
\hat{\bw}&=&(\xi\bSigma+\lambda \bI_p)^{-1}\left(\sqrt{\xi_0}\bSigma^{1/2}\bz+\sqrt{p}R_0\hat{\bmu}\right).
\end{eqnarray}
For $L_1$ penalty $J_\lambda(w)=\lambda|w|$, (\ref{what}) becomes a LASSO type of regression problem which only has a closed form solution in the i.i.d. situation, i.e. $\bSigma=\bI_p$. For general $\bSigma$, we rely on a numerical algorithm to solve $\hat{\bw}$.

\subsection*{Derivation of Corollary \ref{cor}}
The prediction accuracy for binary classification is given by
\begin{eqnarray}\nn
\bE I(y\bx^T\hat{\bw}\ge 0)&=&P(y=1)\bE\{I(\bx^T\hat{\bw}\ge 0)|y=1\}+P(y=-1)\bE\{I(\bx^T\hat{\bw}\le 0)|y=-1\}.
\end{eqnarray}
Conditional on $y=1$, $\bx\sim N(\bmu,\bSigma)$. Thus, $\bx^T\bw\sim N(\bw^T\bmu,\bw^T\bSigma\bw)$ and $\bE\{I(\bx^T\hat{\bw}\ge 0)|y=1\}=\Phi(\delta)$, where $\delta=\frac{\bw^T\bmu}{\sqrt{\bw^T\bSigma\bw}}$. Similar we can derive that conditional on $y=-1$, $\bE\{I(\bx^T\hat{\bw}\le 0)|y=-1\}=\Phi(\delta)$. Therefore $\bE I(y\bx^T\hat{\bw}\ge 0)=\Phi(\delta)$, and from (\ref{eq03}) and (\ref{eq05}), we get $\delta=\frac{R\mu}{\sqrt{q_0}}$.

\bibsep 5pt
\bibliographystyle{chicago}
\bibliography{reference}

\end{document}